# Tightening MRF Relaxations with Planar Subproblems

**Julian Yarkony, Ragib Morshed, Alexander T. Ihler, Charless C. Fowlkes**
Department of Computer Science
University of California, Irvine
{jyarkony,rmorshed,ihler,fowlkes}@ics.uci.edu

## Abstract

We describe a new technique for computing lower-bounds on the minimum energy configuration of a planar Markov Random Field (MRF). Our method successively adds large numbers of constraints and enforces consistency over binary projections of the original problem state space. These constraints are represented in terms of subproblems in a dual-decomposition framework that is optimized using subgradient techniques. The complete set of constraints we consider enforces cycle consistency over the original graph. In practice we find that the method converges quickly on most problems with the addition of a few subproblems and outperforms existing methods for some interesting classes of hard potentials.

## 1 Introduction

A standard approach to finding maximum *a posteriori* (MAP) solutions (or equivalently minimum energy configurations) in a pairwise Markov random field (MRF) is to relax the combinatorial problem to a linear program while enforcing constraints that try to assure integrality of the resulting solution. The chief difficulty is that there are a huge number of possible constraints and only a small subset can possibly be enforced. The best understood case is that of imposing consistency constraints on each pair of variables along an edge. This set of constraints is known as the local polytope. As shown by Wainwright et al. (2005), such a relaxation is directly related to another optimization technique known as dual-decomposition. In particular, the set of pairwise constraints can be derived by considering a lower-bound for the minimum energy constructed by allocating the parameters of the original MRF across a set of overlapping trees that cover every edge of the original graph.

While enforcing consistency over pairs is sufficient to guarantee an integral solution for tree-structured graphs, it may yield non-integral results for more general graphs. In the search for more accurate solutions, one is led to consider higher-order constraints such as cycles. Although explicitly enforcing constraints over all cycles can be done with $O(n^3)$ additional constraints, even this can quickly become impractical in large-scale inference problems. Instead, various authors have proposed using cutting-plane or cycle-repair techniques (Sontag and Jaakkola, 2007; Sontag et al., 2008; Komodakis and Paragios, 2008). These methods first solve the problem with a subset of constraints and then analyze the resulting non-integral solution in order to select additional, violated constraints to be added to the active set. This process is repeated with these additional constraints until an integral solution is found (or the set of possible constraints is exhausted).

In this paper we propose a new class of constraints for general planar MRFs which capture consistency over cycles using tractable binary planar problems as a subroutine. Our approach works directly in the dual-decomposition framework, successively adding subproblems that tighten the lower-bound. Furthermore, these sub-problems are efficient to solve but typically encode many constraints simultaneously.

## 2 Lower-bounds on MRF energy

Consider the problem of minimizing the energy function $E(X)$ associated with an MRF defined over a collection of variables $(X_1, X_2, \ldots, X_N) \in \{1, \ldots, D\}^N$ with specified unary and pairwise potentials. We write

$$E(X, \theta) = \sum_i \theta_i(X_i) + \sum_{i,j} \theta_{ij}(X_i, X_j) \quad (1)$$

where the pairwise and unary functions are described by the collection of parameters $\theta = \{\theta_{i;u}, \theta_{ij;uv}\}$ with

$$\theta_i(X_i) = \sum_u \theta_{i;u} X_{i;u}$$

and

$$\theta_{ij}(X_i, X_j) = \sum_{u,v} \theta_{ij;uv} X_{i;u} X_{j;v}$$

respectively, and $X_{i;u} = [X_i = u]$ denotes a binary indicator that variable $X_i$ takes on state $u$.

Finding a minimum energy configuration is intractable for general $E(X, \theta)$. In this paper we exploit two specific cases in which minima can be found efficiently. Consider the undirected graph $\mathcal{G}(\theta)$ over the $N$ variable nodes that contains edges $(i, j)$ for those $\theta_{ij}$ that are non-zero. If $\mathcal{G}(\theta)$ is a tree then an exact solution can be found in $O(ND^2)$ time using dynamic programming. If $\mathcal{G}(\theta)$ is a planar graph, the problem is binary ($D = 2$) and $E$ includes only symmetric pairwise potentials (i.e., $\theta_{ij}(X_i, X_j)$ only depends on whether a pair of nodes $X_i, X_j$ are in the same or different states) then an exact solution can be found in $O(N^{3/2} \log N)$ by reduction to planar matching (Kasteleyn, 1961; Fisher, 1961; Shih et al., 1990), and an efficient implementation of this technique is available (Schraudolph and Kamenetsky, 2008).

For the planar symmetric case, the energy function can be written more simply as

$$E(X, \vartheta) = \sum_{i>j} \vartheta_{ij} [X_i \neq X_j] + C \quad (2)$$

where $\vartheta_{ij}$ is a scalar that captures the relative energy of $X_i$ and $X_j$ *disagreeing* and $C$ is a constant representing the energy when all nodes take on the same state. Minimizing this energy function can be interpreted as the problem of finding a bi-partition of the graph $\mathcal{G}(\vartheta)$ where the cost of a partition is simply the sum of the weights $\vartheta_{ij}$ of edges cut. Given a minimal weight partition, we can find a corresponding optimal state $X$ by assigning all the nodes in one partition to state 0 and the complement to state 1. Since the edge weights $\vartheta_{ij}$ may be negative, such a minimum weight partition is typically non-empty.

### 2.1 Tree Decomposition and Edge Consistency

To tackle more general pairwise MRFs, we consider the problem of finding a lower-bound on the minimum energy configuration. One promising approach is to decompose $\theta$ into tractable subproblems which each can be solved independently. For example, Wainwright et al. (2005) consider tree subproblems that are subgraphs of the original graph. The decomposition approach is quite appealing as one can consider subproblem decomposition into other structures beyond trees such as planar graphs (Globerson and Jaakkola, 2007a), outer-planar graphs (Batra et al., 2010), k-fans (Kappes et al., 2010), or some heterogeneous mix of subproblems.

Let $t$ index subproblems defined over the same set of variables with parameter vectors $\theta^t$ that sum up to the original parameter vector $\theta = \sum_t \theta^t$. Since the energy function is linear in $\theta$ we can bound the energy of the MAP by

$$\min_X E(X, \theta) = \min_X \sum_t E(X, \theta^t) \quad (3)$$

$$\geq \max_{\substack{\{\theta^t\} \\ \sum_t \theta^t = \theta}} \sum_t \min_{X^t} E(X^t, \theta^t) \quad (4)$$

The inequality arises because the minimization inside the sum may choose different solutions for each subproblem. However, if the solutions $\{X^t\}$ to all the subproblems agree, then the lower-bound is tight. The maximization on the right hand side of Equation 4 is a convex optimization problem with respect to the parameters $\{\theta^t\}$, so if the minimum of each $E(X^t, \theta^t)$ can be found efficiently, then the tightest such bound can be computed using standard constrained subgradient techniques.

Our approach starts with a problem decomposition that is mathematically equivalent to the tree decomposition used in tree-reweighted belief propagation (TRW) (Wainwright et al., 2005; Kolmogorov, 2006). Consider the simplest tree decomposition in which we break the original graph into the collection of trees, each of which consists of a single edge. To accomplish this, a node $X_i$ with degree $d_i$ will be duplicated $d_i$ times, once for each incident edge. We will use $X_i^t$ to refer to the copies of variable $X_i$ in our tree decomposition, where the range of the index $t$ may depend on which variable $X_i$ we are considering. We can bound the MAP energy by

$$E_{MAP} \geq \max_{\substack{\{\theta^t\} \\ \sum_t \theta^t_{i;u} = \theta_{i;u}}} \min_{\{X^t\}} \sum_{i,t} \sum_u \theta^t_{i;u} X^t_{i;u} +$$

$$\sum_{(i,j)} \sum_{u,v} \theta_{ij;uv} X^{t_{ij}}_{i;u} X^{t_{ji}}_{j;v} \quad (5)$$

where $(i, j)$ are the edges in the original graph, and $t_{ij}$ is the copy of node $i$ which is adjacent to a copy of node $j$ in the collection of single edge subproblems.

This lower-bound is the same as the bound given by any decomposition into trees that covers every edge in the original graph. In practice, it is more efficient to consider larger trees, reducing the total number of duplicate nodes in the decomposition. In our experiments, we follow Yarkony et al. (2010) in assembling these edges into a single "covering tree" which has the fewest number of duplicated nodes and covers every edge in the original graph. In this case, the minimization over $\{X^t\}$ is carried out in a single pass of dynamic programming over a tree which has the same edges as the original graph. The duplicate unary parameters $\theta_{i;u}^t$ are then modified using subgradient or fix point updates in order to maximize agreement between duplicate copies of each $X_i$.

A powerful tool for understanding the maximization in Equation 5 is to work with the Lagrangian dual. Equation 5 is an integer linear program over $X$, but the integrality constraints can be relaxed to a linear program over continuous parameters $\mu$ representing min-marginals which are constrained to lie within the *marginal polytope*, $\mu \in \mathbb{M}(\mathcal{G})$. The set of constraints that define $\mathbb{M}(\mathcal{G})$ are a function of the graph structure $\mathcal{G}$ and are defined by an (exponentially large) set of linear constraints that restrict $\mu$ to the set of min-marginals achievable by some consistent joint distribution (see Wainwright and Jordan, 2008). Lower-bounds of the form in Equation 5 correspond to relaxing this set of constraints to the intersection of the constraints enforced by the structure of each subproblem. For the tree-structured subproblems, this relaxation results in the so-called *local* polytope $\mathbb{L}(\mathcal{G})$ which enforces marginalization constraints on each edge. Since $\mathbb{L}(\mathcal{G})$ is an outer bound on $\mathbb{M}(\mathcal{G})$, minimization yields a lower-bound on the original problem. For any relaxed set of constraints, the values of $\mu$ may not correspond to the min-marginals of any valid distribution, and so are referred to as pseudo-marginals.

While enforcing consistency over edges is guaranteed to give an integral solution for tree-structured graphs, it will not yield true min-marginals for more general graphs. In such a case, it is necessary to consider adding higher-order constraints in order to tighten the outer bound. One natural class of higher-order constraints beyond edge consistency is to enforce consistency over cycles of the graph. This can be achieved in various ways. For example, one can triangulate the graph and introduce constraints over all triplets in the resulting triangulation. However, this involves $O(n^3)$ constraints which is impractical in large-scale inference problems. A more efficient route is to only add a small number of constraints as needed, e.g., using a cutting-plane approach (Sontag and Jaakkola, 2007; Sontag et al., 2008; Komodakis and Paragios, 2008). In the following section we propose a new class of subproblems/constraints that enforce consistency over large collections of cycles in planar graphs.

## 2.2 Binary Planar Subproblems and Cycle Consistency

We would like to exploit the tractability of binary planar problems in tightening the lower-bound. There is clearly not a one-to-one mapping from our $D$-state MRF to a binary planar problem in general. However, if the pairwise potentials happened to take on only two values across the $D \times D$ possible states along every edge, then we could first project down to an equivalent binary problem and then lift the solution back to the original state space.

To be precise, for a given subproblem $k$ with nodes $X_i^k$, suppose we partition the state space of the original variables $X_i$ into two subsets $S_i^k \cup \bar{S}_i^k = \{1 \ldots D\}$. We allow for each node $i$ in each subproblem $k$ to have a distinct partition. We will say that the potentials $\theta^k$ are of the *planar binary type* if $\mathcal{G}(\theta^k)$ is planar, $\theta_{i;u}^k = 0$ and $\theta_{ij}^k$ is of the form:

$$\theta_{ij}^k(X_i, X_j) = \begin{cases} \vartheta_{ij}^k & : \quad (X_i^k \in S_i^k) \oplus (X_j^k \in S_j^k) \\ 0 & : \quad \text{otherwise} \end{cases}$$

where $\oplus$ denotes exclusive-or. Then we can define a projected energy function of the form shown in Equation 2 with binary variables $\hat{X}_i^k$ where $\hat{X}_i^k = 1 \iff X_i^k \in S_i^k$ and edge weights $\vartheta_{ij}^k$. The solution to this binary problem can be found efficiently. Furthermore, we can lift it back to a solution with the same energy in the original state space by setting $X_i^k$ to some element of $S_i^k$ when $\hat{X}_i^k = 1$ and to some element of $\bar{S}_i^k$ otherwise. Since the pairwise potentials are constant on these sets, it does not matter which value we choose; all such solutions have the same energy.

It turns out that such binary, planar subproblems have a simple interpretation in terms of the constraints they enforce in the relaxation of the marginal polytope $\mathbb{M}$. It can be shown that the marginal polytope for such a problem is precisely the set of *cycle constraints* on the binary graph, closely connecting the use of planar subproblems and algorithms that enforce or repair cycle consistency.

A particularly simple and appealing subset of binarized problems are problems in which the sets are all of the form $S_i^k = \{u_i\}$ for some single state $u_i$. We refer to these as *one-versus-all* problems and use them exclusively in our implementation.

## 3 Bound Optimization

Our algorithm will optimize a lower-bound consisting of a combination of the tree-structured problem and several binary subproblems. We write this optimiza-

tion problem as:

$$E_{MAP} \geq \max_{\{\theta^t,\theta^0,\theta^k\}} \min_{\{X^t,X^k\}} \sum_{i,t} \sum_u \theta^t_{i;u} X^t_{i;u} +$$
$$\sum_{(i,j)} \sum_{u,v} \theta^0_{ij;uv} X^{t_{ij}}_{i;u} X^{t_{ji}}_{j;v} +$$
$$\sum_{(i,j),k} \sum_{u,v} \theta^k_{ij;uv} X^k_{i;u} X^k_{j;v} \quad (6)$$

subject to the constraints:

$$\sum_t \theta^t_{i;u} = \theta_{i;u}$$
$$\sum_k \theta^k_{ij;uv} + \theta^0_{ij;uv} = \theta_{ij;uv}$$
$\theta^k$ is of the planar binary type

By convention, index $t$ always runs over the copies of nodes $X^t_i$ in the tree structured problem and index $k$ always runs over copies in the planar problems. We use $\theta^0_{ij}$ to denote the allocation of the pairwise potentials to the tree structured subproblem.

In optimizing the bound, we are trying to find an allocation of the unary parameters among copies of the nodes in the tree and allocations of the pairwise parameters across the tree and binary planar subproblems. A straightforward approach to solving this bound optimization problem is to use projected subgradient techniques. For a fixed value of the $X$ variables, the function is linear in $\theta$ with linear equality constraints. In the optimization, one alternates between (1) solving for the $X$ using dynamic programming to find $\{X^t\}$ and perfect matching to find $\{X^k\}$ and (2) updating $\theta$ using a projected-gradient step.

### 3.1 Parameter updates

In the supplemental material we derive the gradient updates and projections for a given setting of the $X$ variables. In this section, we give a concise expression for the complete rule updates which combine both the gradient and projection steps.

We first need some notation to pick out those sets of binary subproblems which are relevant to a given edge state. For a pair of states $(u,v)$ along an edge $(i,j)$, consider the collection of pairs $(s_1, s_2)$ where $s_i$ are subsets of the state values, $s_i \subset \{1, \ldots, D\}$, given by

$$\mathcal{S}_{ij;uv} = \{(s_1, s_2) : (u \in s_1) \oplus (v \in s_2)\};$$

$\mathcal{S}$ indexes all subsets of states in which the configuration $X_i = u, X_j = v$ would produce disagreement. Also, let $Q_{ij;s_1,s_2} = \{k : (s_1 = S^k_i) \wedge (s_2 = S^k_j)\}$ be the set of binary problems containing those subsets for nodes $i$ and $j$ respectively.

Let $\hat{X}^k_{ij} = (X^k_i \in S^k_i) \oplus (X^k_j \in S^k_j)$ be the indicator that in the solution to the $k$th planar subproblem, $X^k_i$ and $X^k_j$ took on disagreeing states. Similarly, let $\hat{X}^0_{ij;s_1s_2} = (X^{t_{ij}}_i \in s_1) \oplus (X^{t_{ji}}_j \in s_2)$ be the indicator that the copies of $X_i$ and $X_j$ which share edge $(i,j)$ in the tree disagreed with respect to the binary projection onto sets $(s_1, s_2)$.

We can then describe the parameter updates compactly in terms of these index sets and indicators

$$\theta^t_{i;u} = \theta^t_{i;u} + \lambda \left( X^t_{i;u} - \frac{\sum_s X^s_{i;u}}{d_i} \right) \quad (7)$$

$$\theta^0_{ij;uv} = \theta^0_{ij;uv} + \lambda \sum_{(s_1,s_2) \in \mathcal{S}_{ij;uv}} \left( \hat{X}^0_{ij,s_1s_2} - \frac{\sum_{\ell \in Q_{ij;s_1s_2}} \hat{X}^\ell_{ij} + \hat{X}^0_{ij,s_1s_2}}{|Q_{ij;s_1s_2}| + 1} \right)$$

$$\vartheta^k_{ij} = \vartheta^k_{ij} + \lambda \left( \hat{X}^k_{ij} - \frac{\sum_{\ell \in Q_{ij;S^k_i S^k_j}} \hat{X}^\ell_{ij} + \hat{X}^0_{ij;S^k_i S^k_j}}{|Q_{ij;S^k_i S^k_j}| + 1} \right)$$

where $d_i$ is the number of duplicate copies of node $i$ in the tree structured problem. Note that the number of sets in $\mathcal{S}$ is small, since $Q_{ij,s_1s_2}$ is empty except for at most $K$ pairs $(s_1, s_2)$, where $K$ is the number of added planar subproblems.

While the notation is rather hairy, these updates have a very intuitive explanation. If the optimal state node in the tree structured problem disagrees with the average of its duplicates, the energy for being in that state is increased and the energy for being in all other states is decreased. If edge $(i,j)$ in the tree takes on state $u, v$ but one or more planar subproblems suggests that either $i \neq u$ or $j \neq v$ then the cost of this state pair along the tree edge is increased. Similarly, if an edge in a subproblem $k$ is cut, then the energy for being cut can be increased by taking away from the covering tree and the other planar subproblems which share the edge parameter.

### 3.2 Repairing Cycles

As there are a huge number of possible binary problems that could be added, we will only use a small set specified by $\{S^1, S^2, \ldots S^P\}$. We suggest the following procedure for successively choosing them:

1. Optimize the current bound until convergence.

2. Find the lowest energy decoding $X$, providing an upper-bound.

3. If the lower-bound and current upper bound are equal then terminate; the current solution is a MAP solution.

4. Otherwise add a new binary subproblem and return to step 1.

In order to compute an upper-bound in step 2, we select a contiguous subtree of the covering tree in which each variable appears exactly once (i.e., a spanning tree of the original graph), and adopt the configuration taken on by those copies. We ensure that such a spanning tree exists by explicitly instantiating it when the covering tree is constructed.

Suppose that $X$ is a decoding for our most recently computed upper-bound found in step 2. To construct a new binary subproblem, we simply specify the collections $S^{P+1}$ by $S_i^{P+1} = \{X_i\}$. In other words, we create a new binary subproblem that checks whether each node $i$ is equal or not equal to its value in the current upper-bound.

When a subproblem is added it restricts the set of inconsistent solutions available to any given cycle, and removes some number of these solutions. However, unlike a true cutting plane technique, it does not guarantee an increase in the lower bound; it removes one or more inconsistent solutions, but cannot ensure that there are no other inconsistent solutions with the same energy. However, we find that in practice this serves as a powerful heuristic. Furthermore, this strategy is also guaranteed to eventually enforce cycle consistency.

**Theorem 3.1** *The procedure of adding binary planar "one-versus-all" subproblems constructed from the most recent upper bound will eventually enforce cycle consistency.*

*Proof Sketch.* Consider any cycle $C$ in the original graph $\mathcal{G}$ and choose a covering tree where the cycle is covered by a chain with a singly copied node $a, a'$. Suppose the parameters have been optimized and that all minimum energy solutions (decodings) for the covering tree, have already been added as planar subproblems. We will show that there must be a minimum energy solution for which the cycle $C$ is consistent.

Suppose that the decoding of cycle $C$ is inconsistent, i.e. there does not exist a minimum energy state of the covering tree in which $a = a'$. Let $X$ be such an inconsistent decoding which uses the state of node $a$ so that the state along the edge including $a'$ is inconsistent with $X$. By assumption, there exists a planar subproblem corresponding to $X$ and the solution along cycle $C$ in the planar subproblem must be consistent with respect to the binary projection (i.e. have an even number of cut edges). In particular the inconsistent edge in the covering tree must be cut in the planar subproblem. However, this contradicts optimality of the bound since the lower-bound could be strictly tightened by increasing the cost of cutting the edge in the planar subproblem. Hence, at convergence there must be no inconsistent cycles. □

An interesting special case is $D = 2$. For binary problems, there is only a single planar binary subproblem to be added and this suffices to enforce all cycle constraints. Furthermore, one can combine the covering tree and planar subproblem into a single larger planar problem containing additional nodes which can be optimized very efficiently. See Yarkony et al. (2011) for details.

Our cycle repair process is generally similar to that of Komodakis and Paragios (2008), which also incrementally repairs violated cycles in a dual decomposition framework. Unlike that work, our cycle repair process adds large collections of cycles (defined by the planar binary graph), giving it the potential to repair many cycles simultaneously.

## 4 Experimental Results

We demonstrate the performance of our algorithm with planar binary subproblems (PSP) on two sets of randomly generated potentials and compare against max-product linear programming (MPLP) (Sontag et al., 2008).

### 4.1 Problem Instances

Our synthetic data consist of two types of problems. For problems of the first type (type-I), each model consists of a grid of size $N \times N$ with $N \in \{10, 20, 30\}$ and $D = 3$ states per node, with random potentials defined as:

$$\theta_{ij;uv} \sim U[-1, 1]$$
$$\theta_{i;u} = 0$$

where $U[-1, 1]$ is a uniform distribution on $[-1, 1]$.

We also generated a second class of problems (type-II), also consisting of grids of size $N \times N$ with $N \in \{10, 20, 30\}$ and 3 states per node, but with potentials described as follows:

$$\theta_{ij;uv} \sim \begin{cases} U[-2, 2] & |u - v| = 1 \\ 0 & u = v \\ 16 & \text{o.w.} \end{cases}$$
$$\theta_{i;u} \sim U[-1, 1]$$

Intuitively, this potential enforces a soft constraint that nodes in state 1 and nodes in state 3 are not neighbors. This type of layered potential occurs in many real-world problems, such as image labeling tasks in computer vision. For example, a scene where water borders the beach, the beach borders the land, but the water cannot border the land. This type of potential is also of theoretical interest since it tends to induce longer loops in the set of necessary cycle constraints.

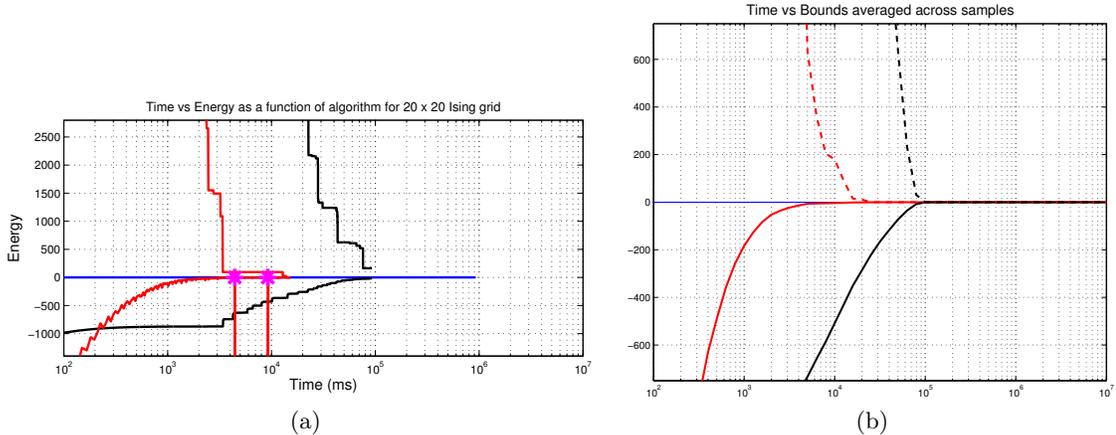

Figure 1: **(a)** Example runs of PSP (red) and MPLP (black) on a single $20 \times 20$ grid problem instance (type-II potentials). Both algorithms continue to repair cycles until the duality gap is closed or the stopping criteria are satisfied. Pink stars indicate the addition of planar subproblems. **(b)** Average convergence behavior across 200 problem instances.

### 4.2 Implementation

We implemented the bound optimization using the BlossomV minimum-weight perfect matching code (Kolmogorov, 2009) and a Bundle Trust subgradient solver (Schramm and Zowe, 1992) to optimize the lower bound. At each iteration, we obtain an upper-bound by extracting a configuration of variables from the covering tree (see Section 3.2). When the lower bound has converged, the most recently generated upper-bound configuration is used to create a new binary subproblem to be added to the collection. In our experiments we terminated the algorithm if it had not managed to find the MAP solution after 10 subproblems had been added.

In our implementation, both types of random potentials are multiplied by 100 and then rounded to the nearest integer, so that a duality gap less than one provides a certificate that the best upper-bound solution does in fact achieve the MAP energy.

We explored initializing the algorithm either with no binary subproblems or "hot starting" with three binary subproblems in which $S_i^k = k$ for every node $i$ and the three states $k = 1, 2, 3$. These subproblems are simple binary projections which capture that a given node is in state $k$ or in state "other". We found that for the randomly generated type-I problems, these initial subproblems did not help overall performance but for type-II problems they decreased the total running time of the algorithm by 10-15x. This is presumably because the type-II problems behave locally like a binary problem and hence the one-versus-other subproblem can immediately enforce consistency on all these local cycles.

### 4.3 Results

We compared performance with the implementation of MPLP (Sontag et al., 2008; Globerson and Jaakkola, 2007b) provided by the authors online. MPLP first runs an optimization corresponding to the tree reweighted lower bound (TRW) then successively tightens the bound by trying to identify cycles whose constraints are significantly violated and adding those sub-problems to the collection. For grids it enumerates and estimates the benefit of adding each square of four variables.

Figure 1(a) plots the upper and lower-bounds as a function of time for each algorithm on a $20 \times 20$ grid problem instance with type-II potentials. Pink markers indicate the time points when new subproblems were added in our algorithm. Figure 1(b) shows convergence behavior averaged over 200 problem instances. Bounds are plotted relative to a MAP energy of 0.

Figure 2 plots the time until the duality gap is closed by our algorithm versus the time for MPLP for each problem instance. In some cases MPLP added all of its feasible subproblems but failed to close the duality gap. Similarly, in some cases, our algorithm (PSP) failed to close the duality gap within its fixed number of subproblems. *Points where MPLP (PSP) found the MAP but the other failed are plotted along the right (resp top) edge of the scatter plot.*

In type-I problems, MPLP tends to optimize its lower bound more quickly than PSP, but PSP manages to close the duality gap for (and thus solve) slightly more problems overall. On type-II problems, however, PSP tends to be faster overall, and manages to find the MAP on significantly more problems than MPLP.

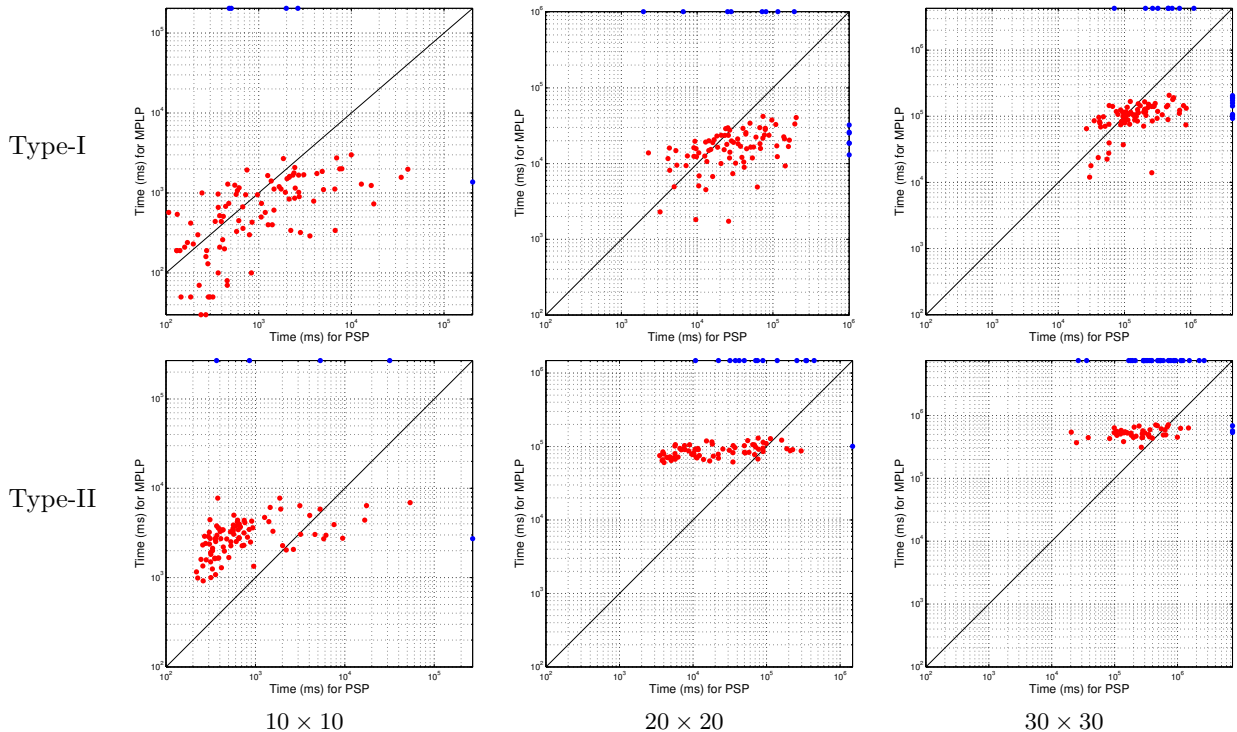

Figure 2: The top row shows problem instances of type-I, and the bottom row from type-II. Points in blue indicate one of the algorithms failed to solve the problem while points in red indicate both algorithms solved the problem. MPLP is slightly faster on type-I data, and PSP slightly faster on type-II; however, PSP solves slightly more type-I problems and significantly more type-II problems than MPLP.

## 5 Discussion

In this work we have described a new variational algorithm for finding minimum energy configurations in planar non-binary MRFs. Our bounds perform incremental cycle repair starting from the tree reweighted (TRW) bounds, adding many cycles at each step through a planar problem construction. Unlike approaches that incrementally add small batches of cycles, we are able to rapidly tighten the bound by adding many potentially violated cycles at once. Our algorithm is competitive with state-of-the-art approaches, with significantly improved performance in models with longer dependency cycles.

#### Acknowledgements

This work was supported by a grant from the UC Labs Research Program and by NSF grant IIS-1065618.

## A Projected Subgradient Updates

Here we derive the projected subgradient updates for a collection of planar subproblems (indexed by $k$) and a single tree-structured problem (indexed by 0). Begin with the unary parameters $\theta_{i;u}$ which are allocated among copies $\theta^t_{i;u}$ that appear in the covering tree. Taking the gradient of the first term in Equation (6) with respect to $\{\theta^t\}$ yields the update rule

$$\theta^t_{i;u} = \theta^t_{i;u} + \lambda X^t_{i;u}$$

However, this neglects the constraint that these parameters must sum up to the original parameter $\sum_t \theta^t_{i;u} = \theta_{i;u}$. We can project the gradient step onto this constraint surface which yields the the valid update

$$\theta^t_{i;u} = \theta^t_{i;u} + \lambda \left( X^t_{i;u} - \frac{\sum_s X^s_{i;u}}{d_i} \right)$$

where $d_i$ is the number of duplicate copies of node $i$. One can easily demonstrate that this update rule preserves the equality constraint by summing both sides of the update equation over all copies $t$.

The updates for the pairwise parameters take a similar form, adjusting each parameter by an amount proportional to the disagreement between its indicator variable and the average of the corresponding indicator variables in other subproblems. However, we have an additional restriction that we can only transfer "disagreement" terms between the tree and the binary planar subproblems.

In order make this explicit let us decompose the pairwise parameters of the tree-structured problem into the original pairwise energies plus a collection of disagreement costs:

$$\theta^0_{ij;uv} = \theta_{ij;uv} + \sum_{(s_1,s_2)\in\mathcal{S}_{ij;uv}} \vartheta^0_{ij,s_1s_2}$$

where $s_1, s_2$ are subsets of the state values, $s_i \subset \{1,\ldots,D\}$, and

$$\mathcal{S}_{ij;uv} = \{(s_1, s_2) : (u \in s_1) \oplus (v \in s_2)\};$$

is the collection of subset pairs in which the configuration $X_i = u, X_j = v$ would produce disagreement. In practice, the only elements of $\mathcal{S}_{ij;uv}$ we care about in the sum are those corresponding to some edge in the planar subproblems we have instanced.

This notation enables us to write the reparameterization constraint succinctly as

$$\vartheta^0_{ij,s_1s_2} + \sum_{k:S^k_i=s_1, S^k_j=s_2} \vartheta^k_{ij} = 0 \qquad \forall\ i,j,s_1,s_2$$

Recall the indicator variables for disagreement along edge $ij$ in our planar subproblem solutions:

$$\hat{X}^k_{ij} = (X^k_i \in S^k_i) \oplus (X^k_j \in S^k_j)$$

We can similarly define a disagreement indicator for the tree-structured problem with respect to some subset pair $(s_1, s_2)$ by

$$\hat{X}^0_{ij,s_1s_2} = (X^{t_{ij}}_i \in s_1) \oplus (X^{t_{ji}}_j \in s_2)$$

In terms of these disagreement indicators, it is easy to show that projected subgradient updates to the planar parameters can be written as

$$\vartheta^k_{ij;S^k_i S^k_j} \leftarrow \vartheta^k_{ij;S^k_i S^k_j} + \lambda \left( \hat{X}^k_{ij} - \frac{\sum_{\ell \in Q_{ij;S^k_i S^k_j}} \hat{X}^\ell_{ij} + \hat{X}^0_{ij,S^k_i S^k_j}}{|Q_{ij;S^k_i S^k_j}| + 1} \right)$$

where

$$Q_{ij;S^k_i S^k_j} = \{\ell\ :\ (S^\ell_i = S^k_i) \wedge (S^\ell_j = S^k_j)\}$$

is the set of planar subproblems (including $k$) that have the same subset pair as $k$ along edge $ij$.

The pairwise parameter updates of the tree affect only $\{\vartheta^0\}$, in a manner analogous to the planar update.

$$\vartheta^0_{ij;s_1s_2} \leftarrow \vartheta^0_{ij;s_1s_2} + \lambda \left( \hat{X}^0_{ij,s_1s_2} - \frac{\sum_{\ell \in Q_{ij;s_1s_2}} \hat{X}^\ell_{ij} + \hat{X}^0_{ij,s_1s_2}}{|Q_{ij;s_1s_2}| + 1} \right).$$

We would prefer to update the $\theta^0$ parameters directly, eliminating the need to keep track of the $\vartheta^0$. Plugging into the definition of $\theta^0$ yields the update from Equation (7) given in the paper

$$\theta^0_{ij;uv} \leftarrow \theta^0_{ij;uv} + \lambda \sum_{(s_1,s_2)\in\mathcal{S}_{ij;uv}} \left( \hat{X}^0_{ij,s_1s_2} - \frac{\sum_{\ell \in Q_{ij;s_1s_2}} \hat{X}^\ell_{ij} + \hat{X}^0_{ij,s_1s_2}}{|Q_{ij;s_1s_2}| + 1} \right).$$